# Network Based Pricing for 3D Printing Services in Two-Sided Manufacturing-as-a-Service Marketplace

Deepak Pahwa, Binil Starly*
Edward. P. Fitts Department of Industrial and Systems Engineering
111 Lampe Drive, North Carolina State University, Raleigh, NC 27695
*Contact author: bstarly@ncsu.edu, Tel: 919 515 1815, Fax: 919 515 5281

**Abstract**

**Purpose**

This paper presents approaches to determine a network based pricing for 3D printing services in the context of a two-sided manufacturing-as-a-service marketplace. The intent is to provide cost analytics to enable service bureaus to better compete in the market by moving away from setting ad-hoc and subjective prices.

**Design/methodology/approach**

A data mining approach with machine learning methods is used to estimate a price range based on the profile characteristics of 3D printing service suppliers. The model considers factors such as supplier experience, supplier capabilities, customer reviews and ratings from past orders, and scale of operations among others to estimate a price range for suppliers' services. Data was gathered from existing marketplace websites, which was then used to train and test the model.

**Findings**

The model demonstrates an accuracy of 65% for US based suppliers and 59% for Europe based suppliers to classify a supplier's 3D Printer listing in one of the seven price categories. The improvement over baseline accuracy of 25% demonstrates that machine learning based methods are promising for network based pricing in manufacturing marketplaces

**Originality/value**

Conventional methodologies for pricing services through activity based costing are inefficient in strategically pricing 3D printing service offering in a connected marketplace. As opposed to arbitrarily determining prices, this work proposes an approach to determine prices through data mining methods to estimate competitive prices. Such tools can be built into online marketplaces to help independent service bureaus to determine service price rates.

## 1. Introduction

The rapid prototyping and short run production service providers now have access to 'Sharing-Economy' type two-sided platforms, which connect designers requesting services to providers who are able to fulfil order requirements (Pahwa et. al, 2018). These platforms have a network of service providers from which they can draw from to satisfy order requirements. Designers in need of 3D Printing services upload designs, receive instant quotes and place orders on these platforms (Rayna and Striukova, 2016). Several approaches have been adopted by these platforms for providing instant pricing services. One approach is to let machine asset owners to set prices which allows designers to select service providers as per their choice. Other platforms utilize advanced heuristics and data mining type algorithms to determine a network based price offering. In the latter case, the platform decides the price for a particular job order. The service provider simply picks job orders to take on, much like ride-sharing platforms.

These platforms centralize the interaction between designers and service providers but are often in control of the prices and the kind of orders the service providers can receive. Ratings, reviews and service type offerings can often lead to a centralized platform driving much of the operations. In decentralized platforms, participants can broadcast services to the entire network of service providers without the need for a 'middle' layer platform. Without a central platform, there is increased transparency, inclusivity and competition for better services. However, decentralized manufacturing systems are often hard to implement, with a major drawback being that price determination by service providers can be ad-hoc and inefficient. Figure 1 presents a graphic summarizing the operation of a two sided 3D printing marketplace.

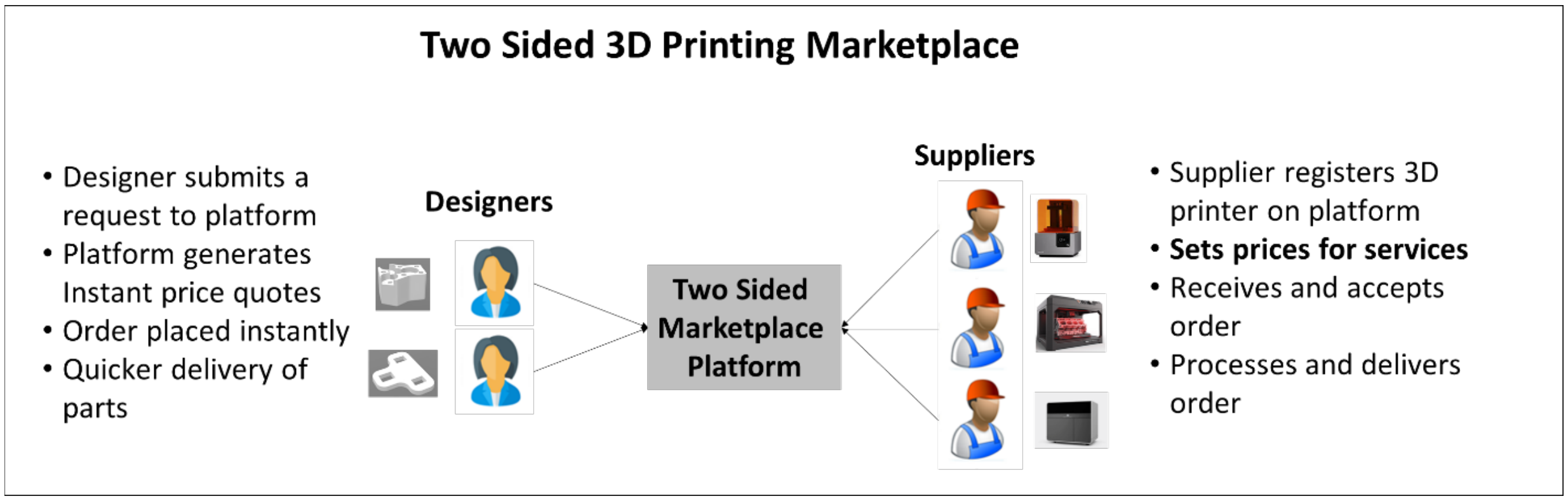

**Figure 1**: Operation of a Two-Sided 3D Printing Marketplace

In an online marketplace, suppliers compete in a national supplier network which makes traditional activity based costing methodologies inefficient to strategically price services within the market. Large service providers have trained professionals, competition benchmarks and other tools to optimally determine prices whereas small suppliers lack the skillsets to determine competitive pricing levels. During the data collection process, price variation among the service providers were significantly high, leading to users having to shop around various service bureaus to get the best price. This extra effort further delays product development times and can discourage designers from prototyping more often.

A naïve approach to competitive pricing for a service provider in an online marketplace would be to compare against features of other suppliers in the neighborhood and then choose a competitive price. Suppliers currently do not systemically link their price to the attributes of their online profile as this manual approach requires significant effort and does not provide precise results. This work seeks to address the question on whether it is possible for small service providers to price services based on specific attributes of their business and services offered through a market driven approach. Publicly available data from the '3D Hubs' marketplace was utilized to build a machine learning model to recommend a price range for services offered by a provider in an online 3D printing marketplace.

## 2. Literature Review

Traditional methods to price additive manufacturing services consist of estimating direct and indirect costs such as raw material and energy costs. One of the first cost models proposed by Hopkinson and Dickens (2003) considers machine costs, labor costs and material costs. They assume negligible energy costs and consider annual production of the same part on one machine. Ruffo et al. (2006) considers a relatively higher impact of the overhead cost of laser sintering technology. Rickenbacher et al. (2013) precisely determines cost per part when parts are printed simultaneously using Selective Laser Melting and Lindemann et al. (2012) considers the lifecycle cost, including the cost of pre and post processes, of an additive manufactured part in its model. Schröder et al. (2015) proposes an activity-based analysis of the process cost and a sensitivity analysis to determine the key cost influencing factors.

Sharing economy platforms in other industries have price recommendation tools to support the participants on the platform. Airbnb has developed a data mining tool (Hill, 2015) which uses features of a listing (among other parameters such as future demand) such as location, amenities, guest reviews to predict a price range of a listing and associated probabilities of success for a sale. Tang and Sangani (2015) predict price category and neighborhood category for Airbnb listings in San Francisco. They use data from 'Inside Airbnb' project (Inside Airbnb Dataset) to create a Machine Learning classifier to make the predictions. Chen and Xie (2017) and Gibbs et al. (2018) study the effect of attributes of a listing such as functionality and reputation on price using a hedonic pricing model for listings from Texas and Canada respectively.

In manufacturing, data-driven approaches for cost estimation of 3D models has been carried out by Chan et. al (2017). Druan et al. (2012) developed artificial neural network based models to estimate the cost of piping elements during early design phase. The data driven methods proposed by these papers do not consider the impact of competition or the complexities associated with a networked service platform. Sharing Economy platforms in additive manufacturing industry can benefit by adopting data driven approaches to support suppliers with price prediction. To the best of authors' knowledge, this work is the first implementation of a machine learning method to price 3D Printing services for an online manufacturing marketplace.

## 3. Methodology

**3.1 Data Collection:** Software was built to scrape data from the public profiles of service bureaus on the '3D Hubs' marketplace platform. As of March 2018, there were 29,554 suppliers listed on the marketplace across the globe. Every supplier has a supplier profile which provides information about its service listings and its reputation on the marketplace. A service listing is a unique combination of a 3D printer, material, resolution and a corresponding price. Table 1 provides 21 supplier attributes, information for which was gathered from their profiles published on the platform. After cleaning the data, 546 suppliers with 5,469 service listings located in the United States and 1,043 suppliers with 9,808 service listings located in Europe remained relevant for the study.

**Table 1**: Listing attributes and their description

| Feature Category | Supplier Feature | Description |
|---|---|---|
| Customer Feedback Features | Average Rating | Service evaluation ratings from customer reviews (1 to 5 with 5 being the best and 1 being the worst). Average Rating is the mean of Print Quality Rating, Speed Rating, Service Rating and Communication Rating |
| | Print Quality Rating | |
| | Speed Rating | |
| | Service Rating | |
| | Communication Rating | |
| | Number of Reviews | Number of customer reviews |
| | Average Response Time | Average time taken to respond to a customer's request |
| Supplier Experience and Scale | Activation Date | Date of supplier's activation on '3D Hubs' |
| | Number of Machines | Number of machines registered on '3D Hubs' |
| | Registered Business | Whether the supplier has a registered business |
| Location | Supplier Location | Country, State, City and GPS coordinates |
| Supplier Description Features | Supplier Description | Business Description of supplier |
| | Print Sample Images | Number of print sample/ other images uploaded on the profile |
| Printer and Material Features | 3D Printer Model | Model of the listed 3D Printer; 3D Printer Cost and 3D Printing Process were derived from the model |
| | Material | 3D Printing Material |
| | Resolution | Print Resolution |
| | Order Completion Time | Number of days required to complete an order |
| Target Feature | Price | Price for a 10 cm tall 3D Printed Marvin Model |

### 3.2 Feature Extraction for Dataset Preparation:

Material, 3D Printer Model, Supplier Location and Supplier Description attributes were not usable in their existing form and had to be quantified/categorized in order to be used in the model as predictor variables.

***Material:*** To consider the material as a predictor variable in the dataset, 346 unique materials listed by these suppliers were categorized in 14 categories. Materials were categorized by its type which included: Acrylonitrile Butadiene Styrene (ABS), Polylactic Acid (PLA), Specialty ABS, Specialty PLA, Polyethylene Terephthalate (PET), Specialty PET, Polycarbonate (PC), Specialty PC, Nylon, Specialty Nylon, Flexible Material (Thermoplastic Elastomer/ Polyurethane), Acrylonitrile Styrene Acrylate (ASA), Metals, Resins, Soluble Material (High Impact Polystyrene, Polyvinyl Alcohol) and Others.

***3D Printer Model:*** To consider the impact of the type of 3D printer on the service price, two parameters were considered: cost of the listed 3D printer and type of 3D printing process. The 3D printing process categories were as follows: Fused Deposition Modelling (FDM), Stereolithography (SLA), Laser Sintering (Selective Laser Sintering for polymers and Direct Metal Laser Sintering for Metals) and Jetting (Material and Binder Jetting). 528 unique 3D printers were categorized, and their cost was gathered from online sources.

***Location:*** To understand the impact of a supplier's location on its price, GPS coordinates of suppliers were used to create geographic clusters. K-means Clustering algorithm was used to generate six supplier clusters in United States and nine supplier clusters in Europe. This algorithm assigns each data point to one of K clusters based on the features (GPS coordinates) that are provided. It assigns the data points to clusters such that the total intra-cluster variation is minimized.

***Supplier Description:*** To understand the impact of supplier description on its price and use it as a predictor variable, supplier description feature vectors were created. 100 Suppliers were randomly selected from the dataset and their text description was used to create a dictionary of keywords. The keywords were segmented in to five categories: Design Services (scanning, modelling), Logistics (turnaround time, pick-up, free shipping), Specialties (such as jeweler, dental, medical), Experience (years of experience, profession, education) and Additional services (services such as finishing, polishing, laser cutting). Supplier description feature vectors include count of number of words belonging to each of the five keyword categories. Suppliers in Europe provided their description in regional languages which were translated to English to create the feature vectors.

### 3.3 Dataset Statistics

***Correlation***: Correlation between numeric listing attributes (Average Rating, Print Quality Rating, Speed Rating, Service Rating, Communication Rating, Number of Reviews, Average Response Time, Activation Date (number of days since activation), Order Completion Time, Resolution, Number of Machines, Number of Print Sample Images, 3D Printer Cost and Price) from Table 1 was measured to determine linear or non-linear monotonic relationship between them. All predictor variables had poor correlation (<40%) with the target variable Price. Correlation between the predictor variables was also determined to remove highly correlated variables from the set of predictor variables. Consumer ratings (Average, Print Quality, Speed, Service and Communication Rating) had significant correlation (>90%) between them. However, all other predictor variables had poor correlation (<40%) between them. Considering the high correlation between the consumer ratings, Print Quality, Speed, Service and Communication Rating, these attributes were dropped from the set of predictor variables. Average Rating and all other predictor variables defined in Table 1 were used in the model.

***Distributions:*** Price distribution for service listings was right skewed in both US and Europe with prices varying from $2.36 – $1956 in the US and $3.75 - $2261.5 in Europe. Majority of the listings (84% in US and 89% in Europe) were FDM printers, significantly lesser (13% in US and 10% in Europe) were SLA and rest SLS and Jetting. In terms of material categories, majority of the listings (54% in US and 57% in Europe) were ABS and PLA materials including specialty formulations. Resins constituted 15% of the listings in US and 10% in Europe, and metals constituted only 0.2% of the listings in both US and Europe. Machine cost also formulated a right skewed distribution with cost varying from $175 to ~$1M. The distributions for these listing attributes suggest that majority of the suppliers own low end machines, use low end materials and sell at lower end of the price spectrum.

### 3.4 Model Formulation

The model was formulated with Price as the target variable and rest of the listing attributes as predictor variables. Average Rating, Number of Reviews, Activation Date (number of days since activation), Average Response Time, Order Completion Time, Number of Machines, 3D Printer Cost, Number of Print Sample Images and Resolution were considered as numeric variables and Registered Business, Supplier Location, 3D Printing Process and Material were considered as categorical variables. Figure 2 represents the model formulation.

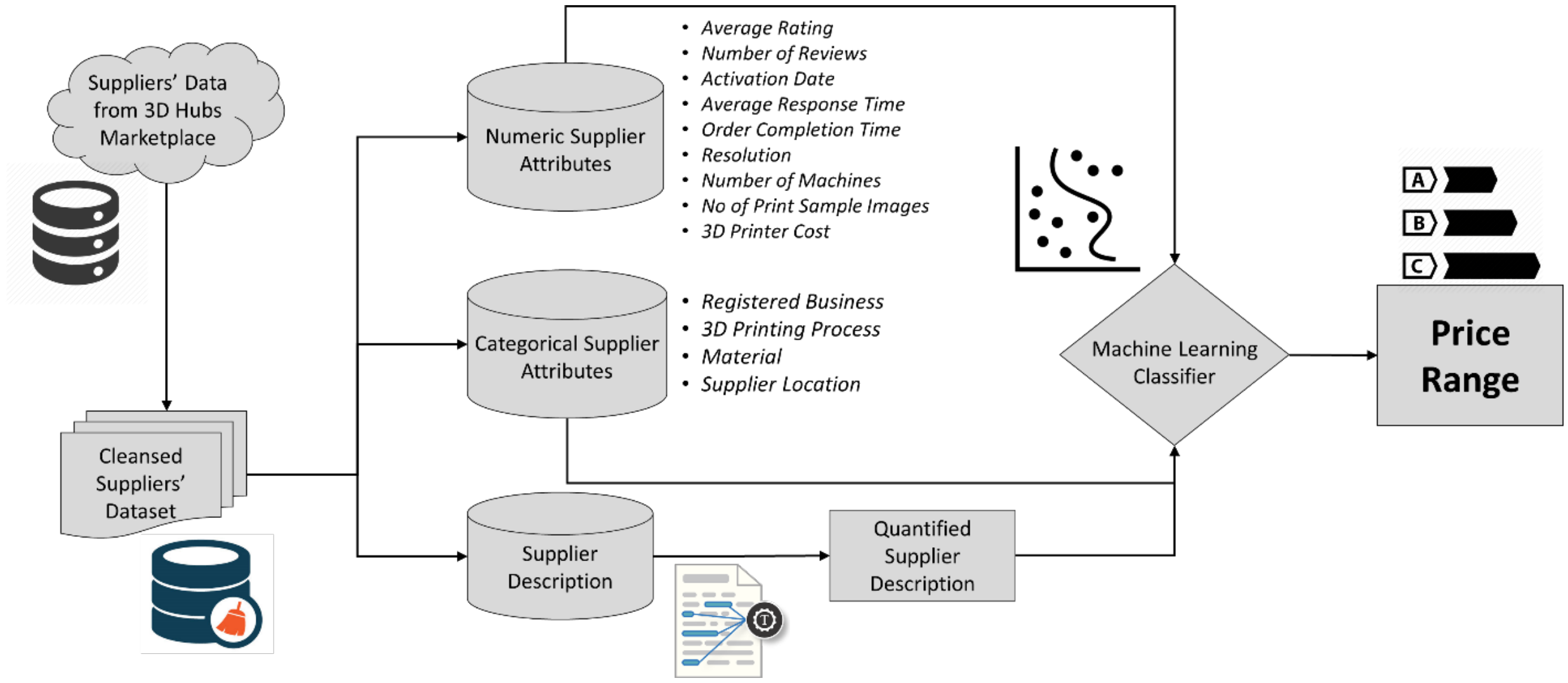

**Figure 2**: Machine Learning Model Formulation

Every supplier on '3D Hubs' listed the price for a 10 cm tall Marvin model for each service listing i.e. unique combination of material, resolution and 3D printer offered by the supplier. Since this was the uniform price metric available for each service listing, it was considered as a price equivalent for service listings' offer. Price was categorized into quartiles resulting in the following four categories in the US: \$2.36 - \$15.1, \$15.1 - \$21.2, \$21.2 - \$36.2, \$36.2 - \$1956. Since the price distribution was right skewed, the price range was extremely wide in the fourth quartile and it was further divided it into quartiles resulting in \$36.2 - \$47.8, \$47.8 - \$64.4, \$64.4 - \$106, \$106 – \$1956. This led to a total of seven price categories. The last category still had a wide price range, however, due to limited observations in this category, it could not be further divided to narrow down the price range.

The data was divided in to training and test sets in an 80:20 ratio using stratified sampling. After preparing the dataset for a classification model, repeated observations were identified in the dataset. For instance, a repeated observation denotes a supplier having two different service listings which result in same listing attributes after material, 3D printer and price categorization. Repeated observations are also important for the dataset as they represent the real-world environment where a supplier can have two FDM printers with a 200-micron resolution ABS material offering with the same price category. However, they can provide misleading results if the same observation is present in both the test and training sets. To overcome this shortcoming, the observations which were present in both test and training set were removed from the test set. The dataset was fit on a Support Vector Machine (SVM) classifier (Chang and Lin, 2011) which solves the following objective function:

$$\min_{w,b,\xi} \frac{1}{2} w^T w + C \sum_{i=1}^{l} \xi_i$$

$$\text{subject to } y_i(w^T \phi(x_i) + b) \geq 1 - \xi_i$$

$$\xi_i \geq 0, i = 1, 2, \ldots\ldots, l$$

$$k(x_i, x_j) = \phi(x_i)^T \phi(x_j)$$

Here $x_i$ represents the vector of listing attributes, $y_i$ represents price classes and $w$ represents the listing attribute weights. $C > 0$ is the regularization parameter and determines the sensitivity of the classifier to misclassification. $\phi(x_i)$ maps $x_i$ to a higher dimension and kernel function $k(x_i, x_j)$ determines the hyperplane which separates the different classes.

## 4. Results and Discussion

The model was trained on the training dataset using a five-fold cross validation and then tested on the test set. Since the classes were imbalanced, the model adjusts the weight of a class inversely proportional to its frequency. The higher the weight of a class, the higher the penalty for misclassification for that class.

***Grid Search:*** To tune the classifier and select optimal hyper parameters, grid search was performed on Kernel (Radial Basis Function (RBF), Linear, Polynomial and Sigmoid kernels), parameters C and γ. Parameter C behaves as a regularization parameter which trades off misclassification against complexity of the model. Parameter γ defines the region of influence of the support vectors selected by the model. Grid search was performed in two steps. First, a relatively larger range of the parameters ($10^{-4}$ to $10^4$) was explored followed by a narrow search around the optimal parameter values found in broader search. RBF Kernel with C = 6500 and γ = 0.01 provided the best results with 72.9% average cross validation accuracy and 65% test set accuracy in the US and 68% average cross validation accuracy and 59.3% test set accuracy in Europe. Grid search provided the optimal hyper parameters to build the classification model.

***Learning Curves:*** Higher values of C can lead to overly complex models resulting in high variance and overfitting. Therefore, to test the model for overfitting, learning curves were plotted. It was found that the model had high variance and was over fitting on the training data with an average training set accuracy of 93% in the US and 92.6% in Europe. To account for overfitting, the model was then tested with lower values of C. C = 100 reduced overfitting significantly with average training accuracy of 78% in the US and 76% in Europe. However, it also reduced both cross validation accuracy and test set accuracy to 65% and 57 % respectively in US and 60% and 53.5% in Europe. Since the model with C = 6500 in US and Europe provided better results on unseen test data, this model was more generalizable and was preferred over models with lower C values.

**Table 2**: Results for SVM Classifier for United States and Europe

| Country | Price Range in $ (Price Class) | Train Accuracy | Cross Validation Accuracy | Test Accuracy | Precision | Recall | F1 Score |
|---|---|---|---|---|---|---|---|
| United States | 2.4 - 15.1 (0) | 0.95 | 0.73 (0.01) | 0.80 | 0.71 | 0.8 | 0.75 |
| | 15.1 - 21.2 (1) | 0.90 | | 0.62 | 0.62 | 0.62 | 0.62 |
| | 21.2 - 36.2 (2) | 0.92 | | 0.61 | 0.67 | 0.61 | 0.64 |
| | 36.2 - 47.8 (3) | 0.99 | | 0.43 | 0.44 | 0.43 | 0.43 |
| | 47.8 - 64.4 (4) | 0.99 | | 0.48 | 0.48 | 0.48 | 0.48 |
| | 64.4 - 106 (5) | 0.99 | | 0.30 | 0.50 | 0.3 | 0.38 |
| | 106 - 1956 (6) | 0.98 | | 0.63 | 0.71 | 0.63 | 0.67 |
| | Micro Average | 0.93 | | 0.65 | 0.65 | 0.65 | 0.65 |
| Europe | 3.8 - 15.9 (0) | 0.94 | 0.68 (0.04) | 0.76 | 0.66 | 0.76 | 0.7 |
| | 15.9 - 21.5 (1) | 0.87 | | 0.54 | 0.56 | 0.54 | 0.55 |
| | 21.5 – 33.1 (2) | 0.91 | | 0.56 | 0.60 | 0.56 | 0.58 |
| | 33.1 - 40.1 (3) | 0.99 | | 0.43 | 0.47 | 0.42 | 0.44 |
| | 40.1 - 56 (4) | 0.99 | | 0.39 | 0.40 | 0.39 | 0.39 |
| | 56 – 87.2(5) | 0.99 | | 0.45 | 0.57 | 0.45 | 0.5 |
| | 87.2 - 2261.5 (6) | 1.00 | | 0.60 | 0.75 | 0.6 | 0.67 |
| | Micro Average | 0.93 | | 0.59 | 0.59 | 0.59 | 0.59 |

Training accuracies, cross validation accuracies on the training set, test set accuracies, precision, recall and F1 score achieved by the model for each price class are presented in Table 2. Cross validation accuracies represents the mean and variance (in parenthesis in Table 2) of accuracies for five folds of cross validation. The baseline accuracy (accuracy with all classes predicted as the class with highest frequency) for this model is 0.25. The micro average test accuracy of 0.65 in US and 0.59 in Europe demonstrates significantly better performance than the baseline.

Precision is a measure of exactness of the classifier i.e. the ratio of correctly predicted positive observations to the total predicted positive observations. For example, for price class 0 in the US, 0.71 precision indicates that 71% of the listings which were predicted to belong to class 0, actually belonged to class 0. Recall is a measure of completeness of the classifier i.e. the ratio of correctly predicted positive observations to the all observations in actual class. For example, for price class 0 in the US, 0.8 recall indicates that 80% of the listings belonging to class 0 were correctly predicted to belong to class 0. F1 score provides a balance between precision and recall and is the harmonic mean of precision and recall. Since multi class models have equal number of false positives and false negatives, the micro averages for test accuracy, precision, recall and F1 score are equal. Therefore, these metrics are more important while evaluating performance of each individual class.

The scores for these metrics are higher for the classes 0, 1 and 2 because each of these classes have higher number of observations (25% each) to learn from. Classes 3, 4 and 5 have significantly lower number of observations (6.25% each) to learn from, leading to lower scores for the scoring metrics. Class 6 shows better performance even with lower number of observations because this class has a wide price range with high end machines and materials. This makes class 6 easily separable from rest of the classes.

Receiver Operator Characteristics (ROC) Curves for US and Europe are presented in Figure 3a and 3b. ROC curves illustrate the performance of a classification model at all classification thresholds and 'Area Under the Curve' (AUC) measures the diagnostic ability of the model. The labels were binarized to plot ROC curves for each price class. Since the classes in the models were imbalanced, the micro average ROC curve which provides an aggregate measure of the performance of the model, was also plotted. 0.89 micro average AUC for the US and 0.87 micro average AUC for Europe represents high predictive accuracy of the classifier to differentiate between the price classes.

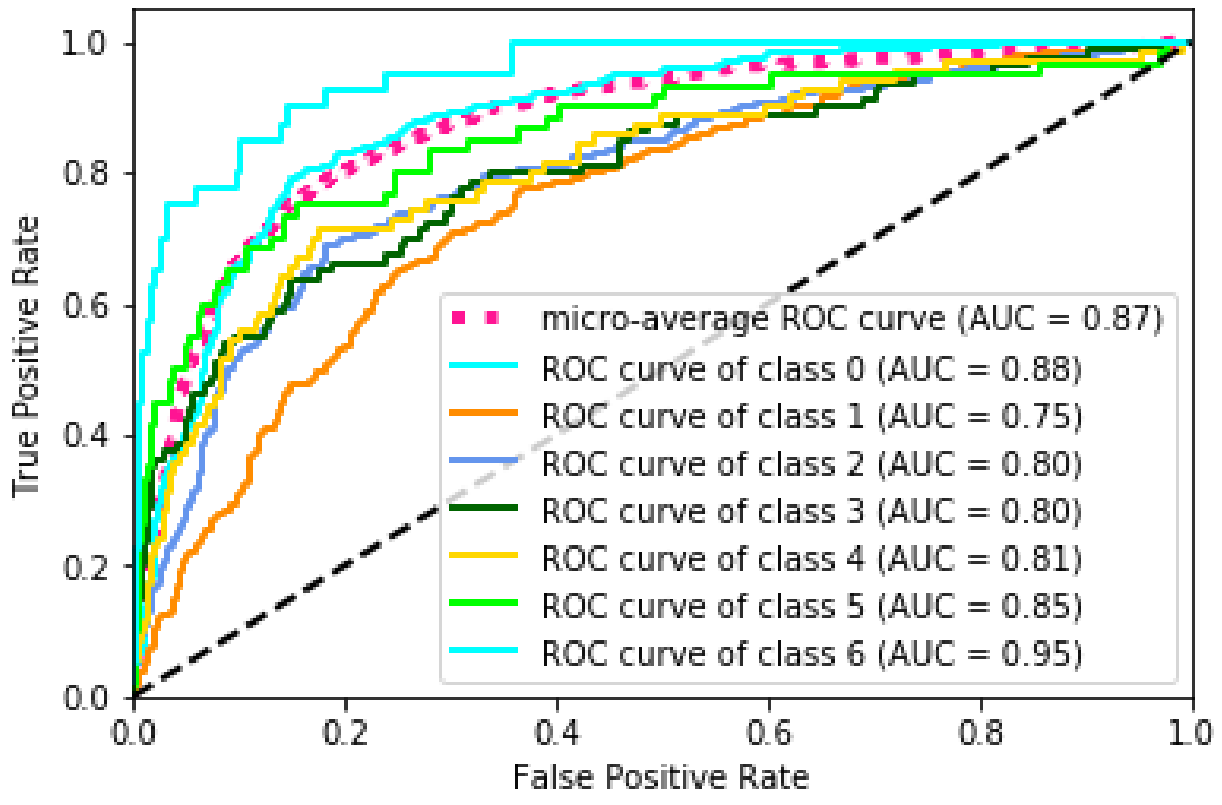

Figure 3a) ROC Curve for Europe

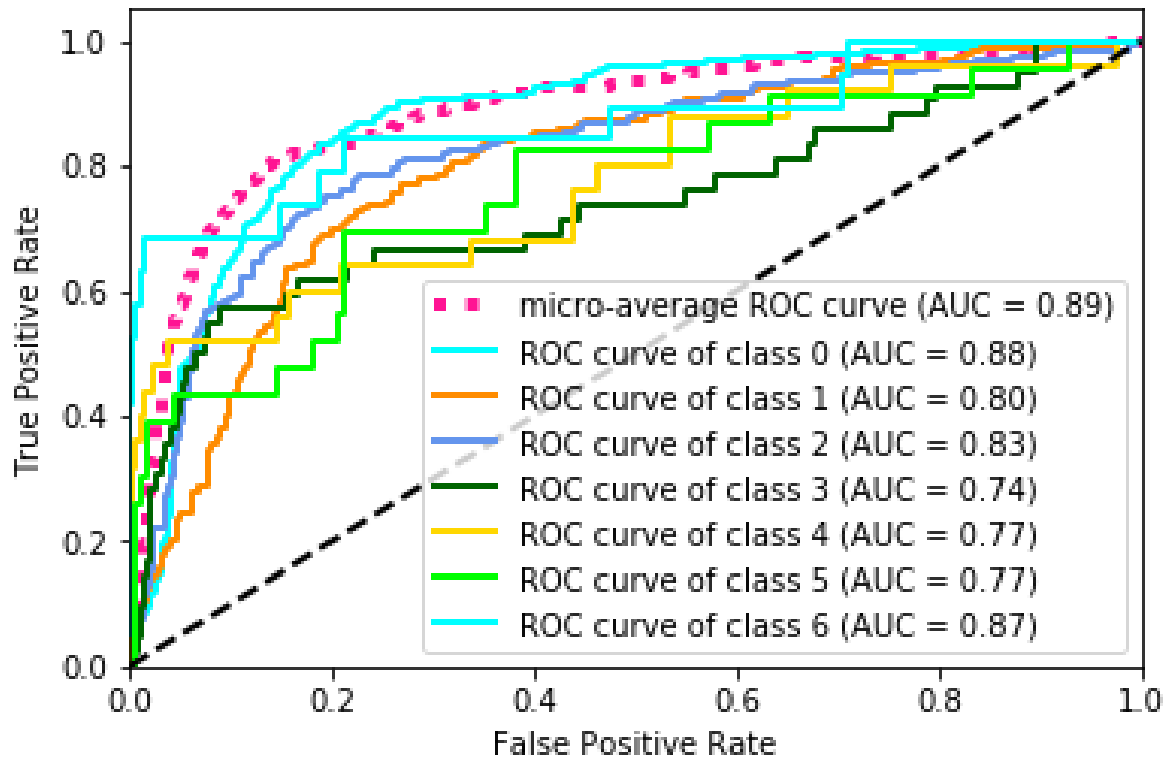

Figure 3b) ROC curve for United States

## 5. Conclusion and Future Work

This work proposes a machine learning approach to determine a network-based price for 3D printing services in an online 3D printing marketplace. The proposed method uses the features of a supplier such as customer reviews and supplier capabilities to predict a price range based on features and prices of other suppliers in the network. The analysis of data from '3D Hubs' marketplace shows that price range of a supplier's listing can be successfully predicted using an array of features extracted from the supplier profiles. The success of data mining-based models for price prediction is only limited by availability of additional data from an online manufacturing marketplace. These marketplaces generate large amount of data which can be used to understand the participants and provide additional value to them. Historical sales data from a manufacturing marketplace could be used to determine the probability of winning an order at a specific price. In addition to supplier's features, order attributes such as 3D design, its due date and designer's attributes such as future potential of orders from the designer can also be considered for price prediction. Impact of additional parameters such as demand forecast, seasonal variation and raw material prices can be added to the model to make it more robust.